\pdfoutput=1

\documentclass[11pt]{article}

\usepackage[]{EMNLP2022}

\usepackage{times}
\usepackage{latexsym}

\usepackage[T1]{fontenc}

\usepackage[utf8]{inputenc}
\usepackage{amsmath}
\usepackage[normalem]{ulem}

\usepackage{hyperref}       
\usepackage{url}            
\usepackage{booktabs}       
\usepackage{amsfonts}       
\usepackage{nicefrac}       
\usepackage[frozencache,cachedir=.]{minted}

\usepackage{tabularx}
\usepackage{multirow}
\usepackage{siunitx}
\usepackage{makecell}
\usepackage{amsmath}
\usepackage{cleveref}
\usepackage{graphicx}

\usepackage{microtype}

\usepackage{inconsolata}

\usepackage{caption}
\usepackage{subcaption}

\usepackage{xspace}
\newcommand{\modelname}{{{\textsc{Asdot}}}\xspace}

\usepackage{color}
\usepackage{colortbl}

\newcommand{\CC}[1]{\cellcolor{lightgray!#1}}
\newcommand{\posimprov}[1]{\footnotesize{(\textcolor{blue}{+#1})}}
\newcommand{\negimprov}[1]{\scriptsize{(\textcolor{Maroon}{-#1})}}

\title{
\modelname: Any-Shot Data-to-Text Generation\\ with Pretrained Language Models}

\author{
Jiannan Xiang$^{1}$,~
Zhengzhong Liu$^{1,3}$,~
Yucheng Zhou$^{2}$,~
Eric P. Xing$^{1,3,4}$,~
Zhiting Hu$^{2}$\\
$^{1}$Carnegie Mellon University,~~$^2$UC San Diego, \\
$^3$Petuum Inc.,~~$^4$Mohamed Bin Zayed University of Artificial Intelligence\\
{\small 
{\tt \{jiannanx,liu,epxing\}@andrew.cmu.edu, \{yuz172,zhh019\}@ucsd.edu}
}
}

\begin{document}
\maketitle
\begin{abstract}
Data-to-text generation is challenging due to the great variety of the input data in terms of domains (e.g., finance vs sports) or schemata (e.g., diverse predicates). Recent end-to-end neural methods thus require substantial training examples to learn to disambiguate and describe the data. 
Yet, real-world data-to-text problems often suffer from various data-scarce issues: one may have access to only a handful of or no training examples, and/or have to rely on examples in a different domain or schema.
To fill this gap, we propose \textit{Any-Shot Data-to-Text} (\modelname{}), a new approach flexibly applicable to diverse settings by making efficient use of any given (or no) examples.
\modelname{} consists of two steps, \textit{data disambiguation} and \textit{sentence fusion}, both of which are amenable to be solved with off-the-shelf pretrained language models (LMs) with optional finetuning.
In the \textit{data disambiguation} stage, we employ the prompted GPT-3 model 
to understand possibly ambiguous triples from the input data and convert each into a short sentence with reduced ambiguity. The \textit{sentence fusion} stage then uses an LM like T5 to fuse all the resulting sentences into a coherent paragraph as the final description. 
We evaluate extensively on various datasets in different scenarios, including the zero-/few-/full-shot settings, and generalization to unseen predicates and out-of-domain data. 
Experimental results show that \modelname{} consistently achieves significant improvement over baselines, e.g., a 30.81 BLEU gain on the DART dataset under the zero-shot setting.\footnote{
{Code available at \url{https://github.com/szxiangjn/any-shot-data2text}}}
\end{abstract}

\section{Introduction}

Data-to-text generation~\cite{kukich1983design,reiter1997building} 
aims at generating natural language text conditioned on structured data content such as tables and graphs. The task has a broad range of applications such as task-oriented dialog~\cite{wen2015semantically}, weather forecasting~\cite{goldberg1994using, sripada2003sumtime}, sports news reporting~\cite{wiseman2017challenges}, and biography generation~\cite{lebret2016neural,wang2018describing}. 


The problem is challenging in practice due to the vast diversity of the input data in terms of the domains (e.g., finance vs sports), schemata (e.g., the set of predicates, table structures), etc. The inherent ambiguity makes it particularly difficult to learn to understand and describe the data.
For instance, in the tuple \texttt{<Fearless, time, 2008>} from a music domain, the predicate word \texttt{time} means the release time of an album, while in \texttt{<100 metres, time, 9.58>} from sports it expresses the world record time. Recent approaches based on end-to-end neural models, e.g., by finetuning pretrained language models (LMs) \citep{puduppully2019data,koncel2019text,zhao2020bridging}, typically require massive training instances to resolve the ambiguity and are not applicable to many data-scarce scenarios. 

In practice, a data-to-text problem of interest may have a varying number of training examples, ranging from a (small) set to only a few shots, or even no examples at all, and sometimes may rely on available examples out of the current domain to facilitate the generation. 
We refer to the diverse practical scenarios as the \emph{any-shot} data-to-text problems. Recent work has studied data-to-text solutions when limited examples are available, but is often restricted to single specific settings. For instance, \newcite{chen-etal-2020-shot} and \newcite{su2021few} focused on few-shot problems but fail to apply when no examples are accessible, while the zero-shot neural pipeline by \newcite{kasner-dusek-2022-neural} relies on human-crafted templates and thus could not handle out-of-domain data.

In this paper, we develop \emph{Any-Shot Data-to-Text} (\modelname{}), a new flexible approach that makes efficient use of any given (or no) examples and achieves stronger generation quality compared to the prior specific methods. \modelname draws inspiration from how humans describe data, namely by first disambiguating and understanding the data content, and then fusing and organizing the information together into text paragraphs. As a result, given input data (e.g., a table or graph), \modelname consists of two intuitive steps, i.e., \emph{data disambiguation} and \emph{sentence fusion}. Importantly, each of the two steps is amenable to be solved with the appropriate off-the-shelf pretrained LMs with optional finetuning, enabling the unique flexibility of \modelname in the presence of any-shot training examples. More specifically, in data disambiguation aiming to understand each data entry (e.g., triple \texttt{<Fearless, time, 2008>}), we use the prompted GPT-3 model~\citep{radford2019language}, which has encoded rich commonsense and world knowledge, to convert the triple into a short sentence (\texttt{Fearless was released in 2008}) with greatly reduced ambiguity. The subsequent sentence fusion stage then uses another LM, such as T5~\citep{raffel2020exploring}, to combine all the resulting sentences into a coherent paragraph as the final description. The sentence fusion as a sub-task allows us to incorporate any available in-/out-of-domain training examples as well as existing large weakly supervised corpus~\cite{kasner-dusek-2022-neural} to finetune the LM and boost the performance.

We evaluate the proposed approach in a wide range of practical any-shot scenarios, including (1) the \emph{zero-/few-/full-shot} setting where we have access to a varying number of training examples, (2) the  \emph{unseen-predicates} setting where we describe the data of new predicates that are never seen in the training examples, and (3) the \emph{out-of-domain} setting where we are presented  only with examples from other domains. Extensive experiments show that our approach consistently achieves significant gains over the diverse previous methods specifically designed for each of the different scenarios.

\vspace{20pt}

\section{Related Work}

Data-to-text (D2T) generation 
is a long-standing problem in natural language processing with broad applications in practice.
Early research on this task focused on rule-based and pipeline approaches~\citep{kukich-1983-design,reiter1997building}, decomposing the task into text planning, sentence planning, and linguistic realisation. Recent work has developed various neural approaches.
\citet{lebret-etal-2016-neural} used a neural encoder-decoder for the task, followed by attention~\citep{bahdanau2015neural}, content selection~\citep{puduppully2019data}, entity modeling~\citep{puduppully-etal-2019-data}, and style imitation~\citep{lin-etal-2020-data} for further improved performance. 
Recent studies have also incorporated pretrained LMs \citep{kale-rastogi-2020-text,ribeiro-etal-2021-investigating,clive2021control}. Although previous fully-supervised methods have achieved remarkable performances, most of them require a large amount of in-domain training examples, leading to limited applicability to the common low-data scenarios in practice. 

Recent interests are aroused in zero-/few-shot data-to-text generation problems. \citet{chen-etal-2020-shot} first formulated the few-shot setting and incorporated a pretrained model with a pointer generator as a solution. \citet{chen-etal-2020-kgpt} developed a knowledge-grounded pretrained LM for both zero- and few-shot data-to-text generation. \citet{gong-etal-2020-tablegpt} and \citet{chen-etal-2020-shot} proposed to solve the few-shot task with content matching and prototype memory, respectively. There are also studies on combining templates and pretrained LM for zero-/few-shot generation. For example, \citet{kale2020template} trained a neural model to rewrite templates for few-shot task-oriented dialogue. \citet{heidari2021getting} applied the idea of template rewriting to build a practical few-shot data-to-text system. 
Most of the previous methods have each focused on a specific setting (e.g., either zero- or few-shot). In comparison, our work studies a wide spectrum of any-shot scenarios with a varying number of training examples from current or different domains. 
Of particular relevance to our work is the approach by \citet{kasner-dusek-2022-neural}, which 
performs zero-shot data-to-text generation by rephrasing given templates. However, the approach relies on human-written templates for data disambiguation and thus has limited applicability to wide domains. Besides, the approach involves several components (ordering, aggregation, compression) to fuse sentences, which restricts the use of any-shot examples for improvement. The approach thus studies only in zero-shot settings, while our work makes a comprehensive study on the diverse any-shot problems.


\begin{figure*}[t]
\centering
  \includegraphics[width=\linewidth]{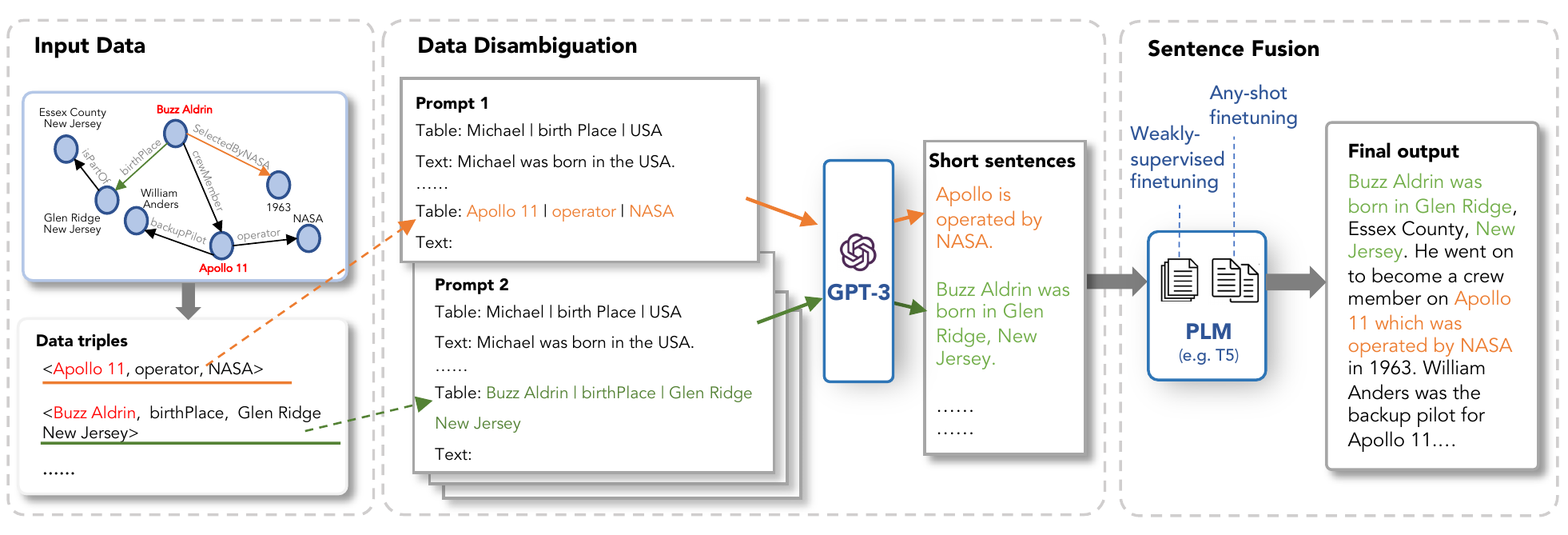}
  \vspace{-20pt}
  \caption{An overview of our method. Our approach consists of two core steps, i.e., \emph{data disambiguation} (\S\ref{sec:tu}) and \emph{sentence fusion} (\S\ref{sec:sf}). The approach first leverages a prompted GPT-3 to convert each data triple into short sentences with reduced ambiguity. The resulting sentences are then fused by a pretrained LM with optional finetuning using public weakly-supervised corpus or available training examples. 
  }
\label{fig:model}
\vspace{-8pt}
\end{figure*}

\section{Any-Shot Data-to-Text Generation}
We propose \modelname{} for any-shot data-to-text generation. \S\ref{sec:setting} describes the any-shot problems. We then provide an overview of our method (\S\ref{sec:method}) and give details of each of the components (\S\ref{sec:tu}, \ref{sec:sf}). Figure~\ref{fig:model} illustrates our method.

\subsection{The Any-Shot Data-to-Text Problems}
\label{sec:setting}
In the data-to-text generation task, we are given structured data (e.g., a table or graph) as input, which can be represented as a set of triples $\{\boldsymbol{x}_1, \boldsymbol{x}_2, ..., \boldsymbol{x}_n\}$. Each triple $\boldsymbol{x}_i = \langle s_i, p_i, o_i \rangle$, such as \texttt{<Apollo 11, operator, NASA>} as in Figure~\ref{fig:model},
consists of a subject $s_i$, a predicate $p_i$, and an object $o_i$, which expresses a relation between the subject and the object. The goal of the task is to generate a paragraph consisting of a sequence of words $\boldsymbol{y} = \{y_1, y_2, ..., y_m\}$ that can describe the input data faithfully and fluently. 

Due to the vast diversity of the content domains, data structures, and predicate sets, etc., building a data-to-text solution often suffers from insufficient training examples for learning to understand/describe the target data. In practice, most often we are presented with a varying number of labeled examples, directly or remotely related to the target data. For instance, we may need to describe a table from a financial report on a new website, where we have no access to any labeled examples (i.e., zero-shot) or have access to only a few description examples (i.e., few-shot). Besides, the available examples may not even be in the financial domain (out of domain), or uses different table structures (different schemata) and different table headers (different predicates). We refer to the data-to-text training in the various practical scenarios as the \textit{any-shot} problem. It is highly desirable to develop a general approach that is widely applicable to the different settings. 


\subsection{Method Overview}\label{sec:method}
Intuitively, a data-to-text generation process consists of two core steps, namely, (1) disambiguating and understanding the data triples, and (2) producing the text description. Previous neural approaches typically model the task in an end-to-end manner and require a large number of training examples to learn the data-to-text mapping. In contrast, we take advantage of the task structure by formulating the two stages and solving each with appropriate resources (e.g., pretrained LMs) that are readily available. Figure~\ref{fig:model} offers an overview of the approach. Specifically, since each data triple is inherently ambiguous given the compact predicate words, rich commonsense and world knowledge is required to correctly understand the content. For instance, in \texttt{<Apollo 11, operator, NASA>}, a model would need knowledge to determine that \texttt{NASA} operates \texttt{Apollo 11} rather than the other way around. Therefore, in the data disambiguation stage, we leverage a powerful LM---GPT-3 in our case---that contains massive implicit knowledge in the parameters, to convert each triple into short sentences with reduced ambiguity (e.g., \texttt{Apollo is operated by NASA}). Once we collect a set of short sentences, in the sentence fusion stage, we use another pretrained LM with optional finetuning to compose the sentences into a well-formed paragraph. The stage offers the flexibility to make use of any available training example to boost performance.


\subsection{Data Disambiguation}\label{sec:tu}
In this stage, the goal is to generate a short sentence to describe each data triple precisely. As above, a triple can be highly abstract and ambiguous as it compresses complex relational information into the compact format $\mathbf{x} = \langle s, p, o \rangle$, where the predicate $p$ is often a concise word or phrase (e.g., the predicate \texttt{time} in triple \texttt{<Fearless, time, 2008>}). To reduce the ambiguity, we want to ``recover'' the missing information in the triple by augmenting it into a complete sentence (e.g., \texttt{Fearless was released in 2008}). Another advantage of converting the structured triples into the free-form text is that a text sequence is more amenable to the LMs used in the subsequent sentence fusion stage (\S\ref{sec:sf}) as described shortly.

As the above examples show, augmenting a triple into a sentence naturally requires relevant external knowledge (e.g., \texttt{Fearless} is an album). 
Training a model specifically for the task could be expensive and could easily overfit to the training domain. Instead, we resort to the general GPT-3 model. Specifically, as shown in Figure~\ref{fig:model} (middle panel), we provide GPT-3 with a few demonstrations of converting triples into short sentences, and then feed the target triple to elicit the desired sentence. Appendix~\ref{sec:prompt} shows the complete demonstrations. We found that the same set of four demonstrations is sufficient to be used for target data in any domain. We thus use the same prompt consisting of those demonstrations throughout our experiments.

Querying the GPT-3 API can be slow and expensive. Given a set of target data in a domain, we reduce the number of queries by generating \emph{templates}. More concretely, for each predicate in the set, we sample one triple containing the predicate, and generate a sentence for the triple with GPT-3. Then we replace the subject and object in the sentence with placeholders \texttt{<subject>} and \texttt{<object>} to get a template. For instance, the template for the predicate \texttt{birthPlace} in Figure~\ref{fig:model} is "\texttt{<subject> was born in <object>}". We then use the template to generate the sentences for all triples with the same predicate.

It is worth noting that many existing data-to-text approaches, ranging from the classical pipeline solutions~\citep{reiter1997building} to the recent neural methods \citep{kale2020template,kasner-dusek-2022-neural}, have also included similar template components, while their templates are typically crafted by human annotators, making the approaches hard to apply to the diverse new domains. In contrast, our \modelname is fully automated with the pretrained LMs, without the need of human efforts nor training examples.

\subsection{Sentence Fusion}\label{sec:sf}
In the second stage, we aim to fuse the sentences from the last step and produce a final coherent and fluent paragraph as the output data description. We naturally formulate the sentence fusion as a sequence-to-sequence problem, and use the pretrained LMs, particularly T5~\citep{raffel2020exploring}, as the backbone for solution. Specifically, we simply concatenate the short sentences, prepended with a prefix word "\texttt{summarize:}", and feed them into the T5 model to obtain the output text. We pick "\texttt{summarize:}" as the prefix for T5 to mimic its pretraining configuration, since the sentence fusion task is similar to the summarization task on which T5 was pretrained. 


\begin{figure*}
\centering
\begin{subfigure}[]{0.45\linewidth}
  \includegraphics[width=\linewidth]{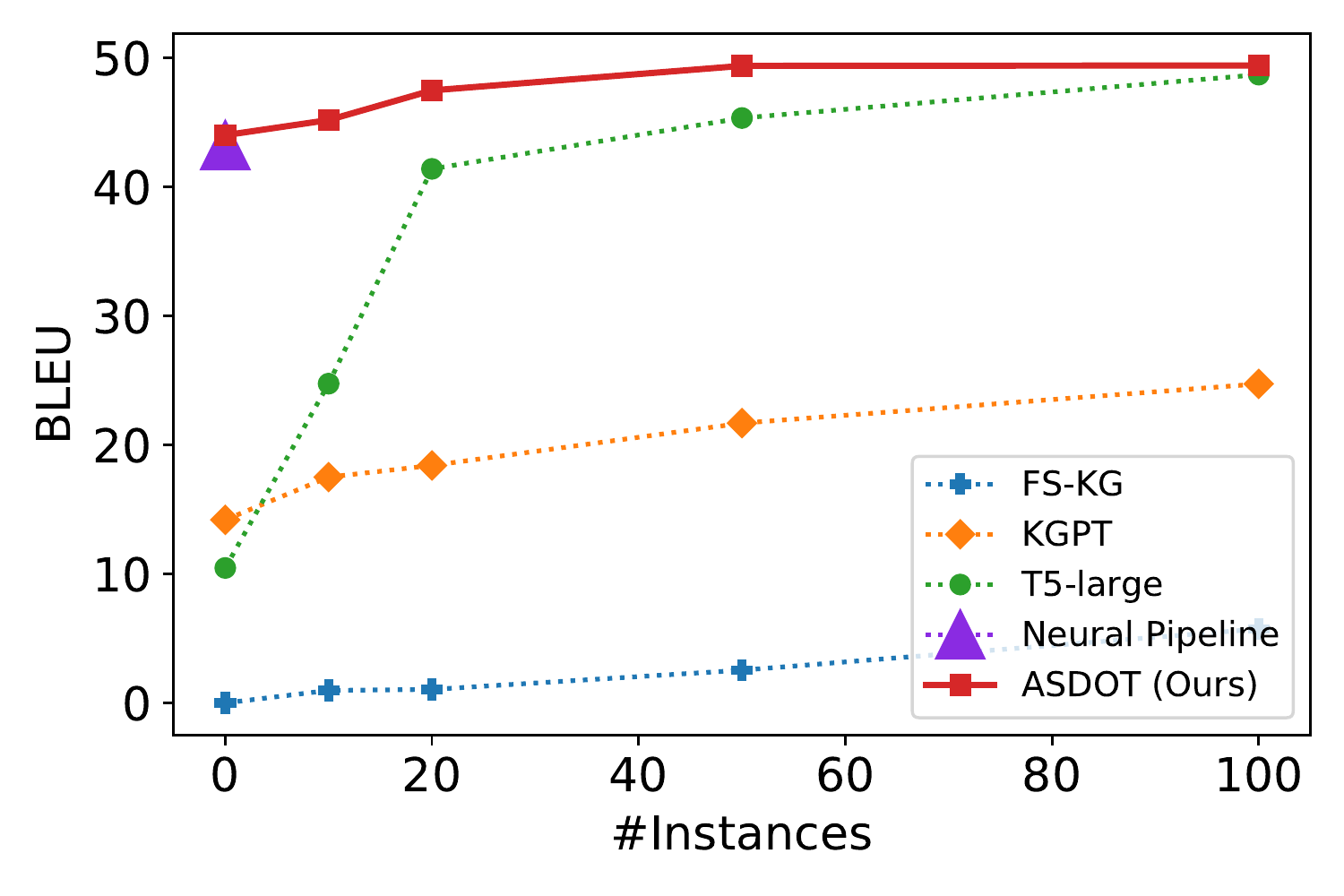}
  \vspace{-2em}
  \caption{WebNLG}
  \end{subfigure}
  \begin{subfigure}[]{0.45\linewidth}
  \includegraphics[width=\linewidth]{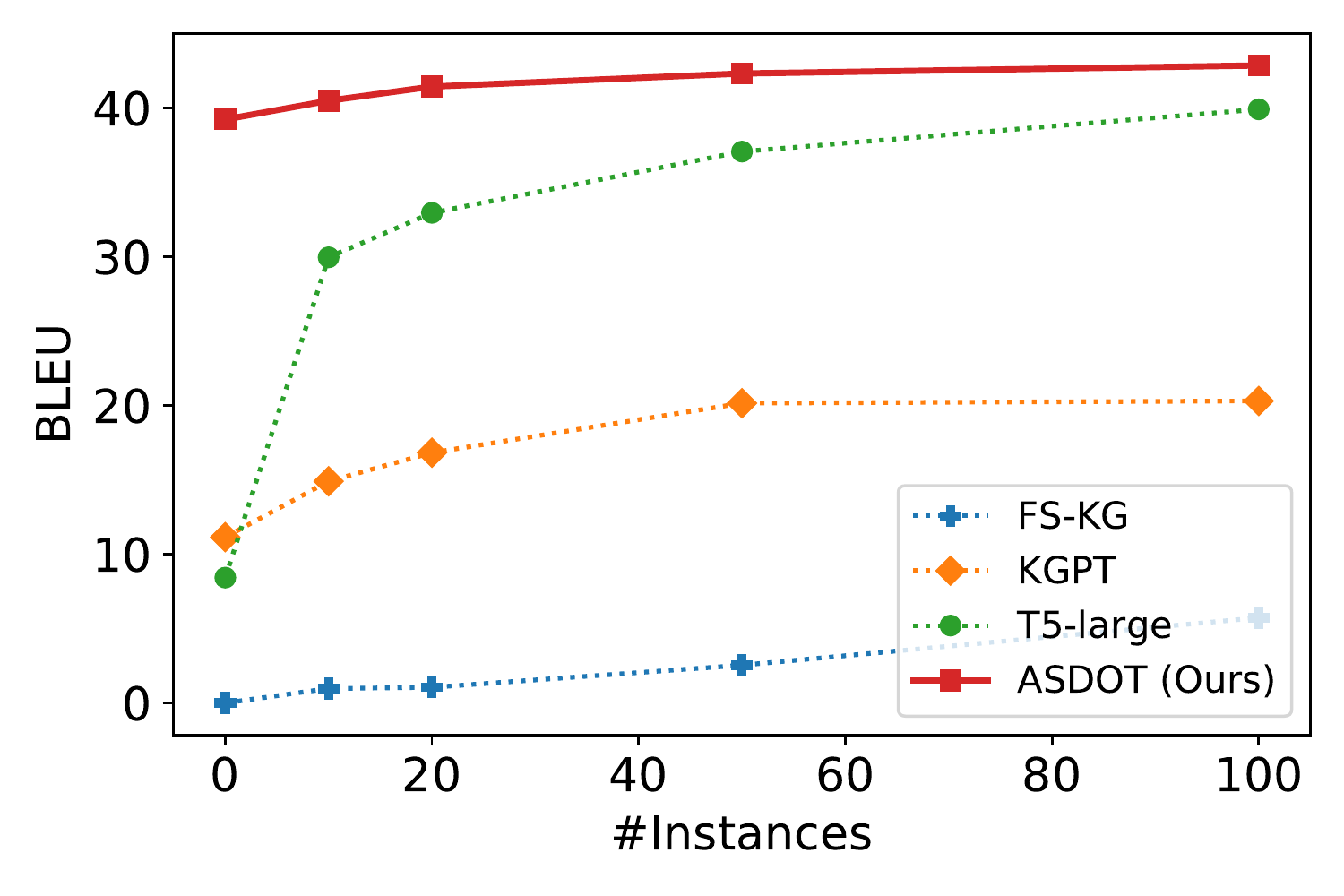}
   \vspace{-2em}
  \caption{DART}
  \end{subfigure}
  \vspace{-8pt}
  \caption{Results of zero-/few-shot learning on WebNLG (left) and DART (right), respectively. The x-axis is the number of training examples, and the y-axis is the BLEU score. We report results of other metrics in Appendix~\ref{sec:appendix any shot}. 
  Neural Pipeline \citep{kasner-dusek-2022-neural} is applicable only to the zero-shot setting and the specific WebNLG data due to the need of human-written templates on the dataset.
  Our method show superior performances under any-shot settings. Our approach shows consistent improvement over the baselines, especially when the training size is small.
  We use paired bootstrap resampling~\citep{koehn-2004-statistical} which confirms that our method is superior to all the baselines at 95\% statistical significance.
  }
\label{fig:anyshot}
\end{figure*} 

A key advantage of the sentence fusion stage is that the component permits easy finetuning with diverse available resources. On one hand, there are automatically constructed weak supervision datasets publicly available, such as WikiSplit \cite{botha2018learning} mined from Wikipedia's edit history and DiscoFuse \cite{geva-etal-2019-discofuse} constructed by rules. In our zero-/few-shot experiments (\S\ref{sec:exp}), we finetune the sentence fusion model with the public WikiFluent dataset~\citep{kasner-dusek-2022-neural} which was constructed by applying a sentence splitting model on the Wikipedia sentences. On the other hand, one can also use any labeled data-to-text examples (by first converting with the data disambiguation stage), even if the examples are from different domains. This is because the general sentence fusion task tends to be domain-agnostic, since the operations to fuse sentences are usually similar across domains, e.g., by inserting connective words or subsuming one sentence as the clause of another. We evaluate in our experiments the out-of-domain generalization ability of our approach.

\section{Experiments}\label{sec:exp}


\subsection{Datasets}\label{sec:dataset}

We experiment on three widely-used data-to-text benchmarks based on which we study various any-shot settings.


{\bf WebNLG} \citep{gardent-etal-2017-creating} consists of data-text pairs where each data is a set of triples extracted from DBpedia and the text is written by human to describe the data. The dataset is split into training, validation, and test set, with  18,102/872/1,862 examples, respectively. The test set is further split into the test-seen and test-unseen subsets. The instances in the test-unseen set are from Wikipedia categories not seen in the training set, which is used in our "unseen predicates" experiments (\S\ref{sec:exp:unseen}).
WebNLG contains 354 types of predicates in total.

{\bf E2E} \citep{novikova-etal-2017-e2e} is a data-to-text corpus in the restaurant domain annotated by human. The dataset has 42,061/547/629 examples in the training/validation/test sets, respectively.
The dataset is relatively easy since it only contains 7 types of predicates and has limited patterns.

{\bf DART} \citep{novikova-etal-2017-e2e} is a large  open-domain data-to-text corpus, constructed from WikiSQL~\citep{zhong2017seq2sql}, WikiTableQuestions~\citep{pasupat-liang-2015-compositional}, as well as the WebNLG and E2E datasets. 
It contains 62,659/2,768/5,097 examples in the training/validation/test sets, respectively, and has 4,299 different predicates in total. 
{Note that the predicates in DART include those in WebNLG and E2E.}
To evaluate model generalization to unseen predicates, we extract a subset of 2,71 test examples whose predicates are completely unseen in the training/validation sets, leading to a more difficult test-unseen set compared to that of WebNLG.

\subsection{Experimental Setup}
For \modelname, the data disambiguation stage (\S\ref{sec:tu}) uses the GPT-3 Davinci API provided by OpenAI,
with greedy decoding, maximum generation length 256 and the stop token "\texttt{\textbackslash n}". Please refer to Appendix~\ref{sec:prompt} for the full prompt we use. 
{As discussed in Section~\ref{sec:tu}, we require only a small number of GPT-3 queries by generating one template for each predicate. Therefore, we query GPT-3 for 4299 times in total, generating for all the predicates in WebNLG, E2E and DART, which costs only \$23 with the GPT-3 pricing as of 10/21/2022.} 
For the sentence fusion stage (\S\ref{sec:sf}), we use T5 models of varying sizes as the sentence fusion LM. In the zero-/few-shot settings (\S\ref{sec:exp:few-shot}), we finetune the T5 with the large weakly-supervised data WikiFluent \citep{kasner-dusek-2022-neural} as mentioned in \S\ref{sec:sf}. We use the Adam optimizer~\citep{kingma2015adam} with an initial learning rate of $3 \times 10^{-5}$, and use a batch size of 64, for 1 epoch.
When any shot of labeled data-to-text examples are available, we further finetune the sentence fusion T5 with those examples.
For the generation, we use beam search decoding with a beam width of 5. We provide more details of the experimental setup in the Appendix~\ref{sec:prompt}.

\begin{table*}[t]
\centering
\small
\begin{minipage}{0.48\textwidth}
\centering
\begin{tabular}{lccc}
	\toprule
    \textbf{Model} &  \textbf{BLEU}& \textbf{METEOR} &\textbf{P-F1} \\ 
    \toprule
    BestPlan &47.24 &39.00 &- \\
    Pipeline-Trans &51.68 &32.00 &- \\
    PlanEnc &52.78 &41.00 &- \\
    DataTuner\_FC &52.40 &42.40 &- \\
    \midrule
     T5-small  &56.90 &43.05 &65.20\\
     \CC{20} &\CC{20}58.64 &\CC{20}43.47 &\CC{20}66.63\\
     \CC{20}\multirow{-2}{*}{\modelname-small}&\CC{20}\posimprov{1.74} &\CC{20}\posimprov{0.42} &\CC{20}\posimprov{1.33}\\
     \midrule
     T5-base &58.53 &43.89 &66.82\\
     \CC{20} &\CC{20}60.34 &\CC{20}44.37 &\CC{20}68.17\\
     \CC{20}\multirow{-2}{*}{\modelname-base} &\CC{20}\posimprov{1.81} &\CC{20}\posimprov{0.48} &\CC{20}\posimprov{1.35}\\
     \midrule
     T5-large &60.38 &44.49 &68.49\\
     \CC{20} &\CC{20}61.32 &\CC{20}\textbf{44.79} &\CC{20}\textbf{69.69}\\
     \CC{20}\multirow{-2}{*}{\modelname-large} &\CC{20}\posimprov{0.94} &\CC{20}\posimprov{0.30} &\CC{20}\posimprov{1.20}\\
     \midrule
    Prefix-Tuning &61.03 &44.37 &69.17\\
    \CC{20} &\CC{20}\textbf{61.38} &\CC{20}44.52 &\CC{20}69.39\\
    \CC{20}\multirow{-2}{*}{\modelname-Prefix} &\CC{20}\posimprov{0.35} &\CC{20}\posimprov{0.15} &\CC{20}\posimprov{0.22}\\
    
	\bottomrule
	\end{tabular}
	\vspace{-6pt}
	\end{minipage}
\begin{minipage}{0.48\textwidth}
\centering
\begin{tabular}{lccc}
	\toprule
	\textbf{Model} &  \textbf{BLEU}& \textbf{METEOR} & \textbf{P-F1}\\ 
    \midrule 
    LSTM w attention &29.66 &27.00 &35.00\\
    E2E Transformer &27.24 &25.00 &28.00\\
    BART-base &47.11 &38.00 &55.00\\
    BART-large &48.56 &39.00 &57.00\\
    \midrule
    {T5-small}  &47.53 &39.00 &59.33\\
    \CC{20} &\CC{20}49.32 &\CC{20}39.57 &\CC{20}60.95\\
    \CC{20}\multirow{-2}{*}{\modelname-small} &\CC{20}\posimprov{1.79} &\CC{20}\posimprov{0.57} &\CC{20}\posimprov{1.62}\\
    \midrule
    
    {T5-base} & 49.62 &39.69 &61.11\\
    \CC{20} &\CC{20}{49.85} &\CC{20}{39.91} &\CC{20}{61.64} \\
    \CC{20}\multirow{-2}{*}{\modelname-base} &\CC{20}\posimprov{0.23} &\CC{20}\posimprov{0.22} &\CC{20}\posimprov{0.53}\\
    \midrule
    
	{T5-large} & 50.17 &40.00 &61.72 \\
	 \CC{20}& \CC{20}\textbf{50.79}   & \CC{20}\textbf{40.36} & \CC{20} \textbf{62.52}\\
	 \CC{20}\multirow{-2}{*}{\modelname-large} &\CC{20}\posimprov{0.62} &\CC{20}\posimprov{0.36} &\CC{20}\posimprov{0.80}\\
	 
	 \midrule
    
	{Prefix-tuning} & 50.39 &40.13 &61.60 \\
	\CC{20} &\CC{20}{50.56}   &\CC{20}{40.22} &\CC{20}{62.27}\\
	\CC{20}\multirow{-2}{*}{\modelname-Prefix} &\CC{20}\posimprov{0.17} &\CC{20}\posimprov{0.09} &\CC{20}\posimprov{0.67}\\
	\bottomrule
	\end{tabular}
	\vspace{-6pt}
	\end{minipage}
	\caption{
	{Full-shot learning results on WebNLG ({\bf Left}) and DART ({\bf Right}). \modelname-{\it X} denotes our approach with T5-{\it X} as the sentence fusion model. The best scores are in {\bf bold}. We also show the performance gains against respective baseline models in} {\color{blue} blue}.}
	\label{table:full-shot}
\end{table*}

\paragraph{Evaluation Metrics}
Following previous studies, we report the performance in terms of BLEU
\citep{papineni-etal-2002-bleu} and METEOR
\citep{banerjee-lavie-2005-meteor}, as well as the recent PARENT-F1 metric \citep{dhingra-etal-2019-handling} which measures the alignment between generated text with both the references and input data. We also report two embedding-based metrics BERTScore~\citep{zhang2019bertscore} and BLEURT~\citep{sellam2020bleurt} in the Appendix~\ref{sec:appendix any shot}. Besides, we perform human evaluation in the few-shot setting as detailed later.




\subsection{Zero-, Few-, to Full-Shot Learning}\label{sec:exp:few-shot}
We evaluate \modelname in the presence of a varying number of training examples, ranging from 0, 10, 20, 50, 100 to the size of the full training set. 
We experiment on the WebNLG and DART datasets, respectively.
In the zero-/few-shot settings, we use the T5-large model for our sentence fusion LM. In the full-shot setting, we test three T5 models of different sizes (small - 60M parameters, base - 220M, and large - 770M) for sentence fusion. Besides, the recent Prefix-Tuning method~\citep{li2021prefix} shows competitive performances on the data-to-text generation task. We thus also incorporate it with the T5-large architecture and report the results.

\paragraph{Baselines}
In the zero-/few-shot settings,
we compare with {\bf KGPT}~\citep{chen-etal-2020-kgpt}, a knowledge-grounded LM pretrained on large-scale automatically constructed data-to-text corpus, as it is one of the few methods applicable to both zero-/few-shot data-to-text generation. 
{Besides, we compare with {\bf FS-KG} \citep{li-etal-2021-shot-knowledge}, a recent few-shot data-to-text approach enhanced with representation alignment between knowledge graphs and PLMs}. We also compare with the end-to-end model based on {\bf T5-large}, which has shown remarkable performance on data-to-text tasks with sufficient training examples~\citep{ribeiro2020investigating}. 
Following \citet{ribeiro-etal-2021-investigating}, for the T5 baseline, we prepend \texttt{<H>}, \texttt{<R>} and \texttt{<T>} before the subjects, predicates, and objects, respectively, and add a prefix "\texttt{translate Graph to English:}" to the input. We finetune the T5 model with available shots of training examples.
On the WebNLG dataset, we report another baseline {\bf Neural Pipeline}~\citep{kasner-dusek-2022-neural}, which is a template-based pipeline method also trained on the WikiFluent dataset and is applicable only to the zero-shot setting. However, the method cannot be used on the DART dataset since its templates are specifically written for WebNLG by human. 

In the full-shot setting, we further compare with a wide range of previous full-shot state-of-the-art data-to-text systems, including BestPlan~\citep{moryossef-etal-2019-step}, Pipeline-Trans~\citep{castro-ferreira-etal-2019-neural}, PlanEnc~\citep{zhao2020bridging}, DataTuner\_FC~\citep{harkous-etal-2020-text} on WebNLG, and LSTM-with-attention, End-to-End Transformers, and BART-base/large~\citep{nan2020dart} on DART. 

\paragraph{Automatic Evaluation}
The zero-/few-shot results are shown in Figure~\ref{fig:anyshot}. Our method consistently outperforms baseline models on both datasets, demonstrating its strong zero-/few-shot learning ability. 
In particular, with fewer training examples, our \modelname tends to outperform other methods by a larger margin. For instance, we achieve 16.06 higher BLEU than T5-large on 10-shot WebNLG, and 10.53 higher on 10-shot DART. This is because the two-stage \modelname is designed to excel in the low-data contexts by augmenting the generation process with rich external knowledge in pretrained LMs.
Neural Pipeline is competitive with ours, but is restricted only to the zero-shot setting on WebNLG.
DART contains more diverse types of predicates and thus is arguably more challenging than WebNLG. Our approach tends to achieve stronger performance gains on the difficult dataset.

We report the results of the full-shot setting in Table~\ref{table:full-shot}. The performance gain tends to be less significant compared to the zero-/few-shot settings as all methods are presented with a large number of training examples.
However, our method still achieves consistently stronger performance over the large diversity of baselines, thanks to \modelname's proper modeling of the generation process and the incorporation of rich external implicit knowledge.  
\begin{table}[t]\centering
\small
\resizebox{1\columnwidth}{!}{
\begin{tabular}{lccc}
	\toprule
	\textbf{Model} & Faithfulness $\uparrow$ & Contradict $\downarrow$& Fluency $\uparrow$\\ 
    \midrule 
    KGPT &0.64 &2.34 &1.00 \\
    T5-large &2.22 &0.72 &1.58 \\
    \midrule
    \modelname & \textbf{2.37}  & \textbf{0.67}  & \textbf{1.82}\\
	\bottomrule
	\end{tabular}}
	\vspace{-8pt}
	\caption{Human evaluation results. $\uparrow$ means the higher the better and $\downarrow$ means the lower the better. \modelname outperforms the baselines with $p<0.05$ in Tukey’s HSD test for all the measures.}
\label{table:human eval}
\end{table}

\paragraph{Human Evaluation} 
We conduct a human evaluation to further assess our \modelname against other baselines under the 50-shot setting on WebNLG. 
After training, we sample 50 test instances and ask three proficient English speakers 
{in the university} to score the model outputs. Following \citet{chen-etal-2020-shot}, each generated result is evaluated on three aspects: the number of the facts that are consistent with the input table (\emph{Faithfulness}) and contradicted to the table (\emph{Contradict}), and the language fluency, on a 3-Likert scale (0,1,2). The results are shown in Table~\ref{table:human eval}. The Krippendorff alphas~\citep{krippendorff1computing} for Faithfulness, Contradict, and language fluency are 0.49, 0.42 and 0.36, respectively, indicating a fair inner-annotator agreement. Consistent with the automatic evaluation results, we observe that \modelname{} is substantially better than the baselines on all the three aspects, suggesting that our approach generates more faithful and fluent descriptions.

\paragraph{Ablation Studies} 
We conduct ablation studies to investigate the effects of both the data disambiguation and sentence fusion stages. Table~\ref{table:ablation} shows the results. Specifically, for the sentence fusion stage, we study the effect of the weakly-supervised finetuning on the WikiFluent corpus (\S\ref{sec:sf}). From the table, we can see that the performance drops sharply without weakly-supervised finetuning, i.e., by 8.86 BLEU points for the zero-shot setting. However, \modelname without weak supervision still outperforms the baselines in most cases, validating the strong advantage of our approach under low-data settings. 
For the data disambiguation stage, we investigate the impact of the automatic templates produced by GPT-3. More concretely, we replace the GPT-3 templates with the human-written templates from \citet{kasner-dusek-2022-neural}. The performance is similar or decreases slightly, demonstrating that the short sentences or templates automatically generated in the data disambiguation stage are of competitive or slightly higher quality than the manually created ones (perhaps due to human errors when writing the hundreds of templates).

\begin{table}[t]\centering
\small
\resizebox{1\columnwidth}{!}{
\begin{tabular}{lccccc}
	\toprule
	\textbf{Model} &  0& 10 &20 &50 &100\\ 
    \midrule 
    KGPT &14.19 &17.50 &18.40 &21.68 &24.72\\
    T5-large &10.46 &29.10 &41.38 &46.24 &48.68\\
    \midrule
    \modelname &\textbf{43.33} &\textbf{45.16} &\textbf{47.46} &\textbf{49.36} &\textbf{49.39} \\
     \ - \emph{w/o weak-sup} &34.47 &39.38 &43.67 &47.56 &48.16\\
     \ - \emph{w/ manual templ.} &42.02 &43.37 &46.12 &48.28 &48.32 \\
	\bottomrule
	\end{tabular}}
	\vspace{-8pt}
	\caption{Ablation results (BLEU) for zero-/few-shot learning on WebNLG. The \emph{w/o weak-sup} row shows the results of \modelname without weakly supervised finetuning, and \emph{w/ manual templ.} shows the results of using hand-crafted templates in the data disambiguation stage.}
\label{table:ablation}
\end{table}

\subsection{Generating for Unseen Predicates}\label{sec:exp:unseen}

\begin{table}
\centering
\small
\begin{tabular}{lccc}
	\toprule
    \textbf{Model} &  \textbf{BLEU}& \textbf{METEOR} &\textbf{P-F1} \\ 
    \midrule
    BestPlan &34.41 &37.00 &- \\
    Pipeline-Trans &38.92 &21.00 &- \\
    PlanEnc &38.23 &37.00 &- \\
    \midrule
     T5-small  &47.34 &39.95 &57.99\\
     \CC{20} &\CC{20}50.75 &\CC{20}40.63 &\CC{20}61.20\\
     \CC{20}\multirow{-2}{*}{\textsc{Asdot}-small}&\CC{20}\posimprov{3.41} &\CC{20}\posimprov{0.68} &\CC{20}\posimprov{3.21}\\
     \midrule
     T5-base &51.11 &41.42 &60.94\\
     \CC{20} &\CC{20}54.51 &\CC{20}42.30 &\CC{20}64.36\\
     \CC{20}\multirow{-2}{*}{\textsc{Asdot}-base} &\CC{20}\posimprov{3.40} &\CC{20}\posimprov{0.88} &\CC{20}\posimprov{3.42}\\
     \midrule
     T5-large &53.97 &42.37 &63.81\\
     \CC{20} &\CC{20}55.74 &\CC{20}\textbf{42.94} &\CC{20}\textbf{65.90}\\
     \CC{20}\multirow{-2}{*}{\textsc{Asdot}-large} &\CC{20}\posimprov{1.77} &\CC{20}\posimprov{0.57} &\CC{20}\posimprov{2.09}\\
     \midrule
    Prefix-Tuning &55.26 &42.42 &65.24\\
    \CC{20} &\CC{20}\textbf{55.86} &\CC{20}42.73 &\CC{20}65.68\\
    \CC{20}\multirow{-2}{*}{\textsc{Asdot}-Prefix} &\CC{20}\posimprov{0.60} &\CC{20}\posimprov{0.31} &\CC{20}\posimprov{0.44}\\
    
	\bottomrule
	\end{tabular}
	\vspace{-8pt}
	\caption{Results on WebNLG test-unseen set.}
	\label{table:webnlg_unseen}
\end{table}

\begin{table}\centering
	\small
\centering
\begin{tabular}{lccc}
	\toprule
	\textbf{Model} &  \textbf{BLEU} & \textbf{METEOR} & \textbf{P-F1}\\ 
    \midrule
    {T5-small}  &37.65 &33.27 &43.79\\
    \CC{20} &\CC{20}46.60 &\CC{20}36.91 &\CC{20}52.17\\
    \CC{20}\multirow{-2}{*}{\textsc{Asdot}-small} &\CC{20}\posimprov{8.95} &\CC{20}\posimprov{3.64} &\CC{20}\posimprov{8.38}\\
    \midrule
    
    {T5-base} & 46.13 & 36.97 &49.79\\
    \CC{20} &\CC{20}{50.90} &\CC{20}{37.72} &\CC{20}{54.98} \\
    \CC{20}\multirow{-2}{*}{\textsc{Asdot}-base} &\CC{20}\posimprov{4.77} &\CC{20}\posimprov{0.75} &\CC{20}\posimprov{5.19}\\
    \midrule
    
	{T5-large} & 46.37  &  36.49 &50.32 \\
	 \CC{20}& \CC{20}{50.70}   & \CC{20}{37.25} & \CC{20} {55.49}\\
	 \CC{20}\multirow{-2}{*}{\textsc{Asdot}-large} &\CC{20}\posimprov{4.33} &\CC{20}\posimprov{0.76} &\CC{20}\posimprov{5.17}\\
	 
	 \midrule
    
	{Prefix-tuning} & 47.07  &  36.69 &49.67 \\
	\CC{20} &\CC{20}\textbf{51.99}   &\CC{20}\textbf{38.11} &\CC{20}\textbf{57.26}\\
	\CC{20}\multirow{-2}{*}{\textsc{Asdot}-Prefix} &\CC{20}\posimprov{4.92} &\CC{20}\posimprov{1.42} &\CC{20}\posimprov{7.59}\\
	\bottomrule
	\end{tabular}
	\vspace{-8pt}
	\caption{Results on DART test-unseen set.}
	\label{table:dart_unseen}
\label{table:unseen}
\vspace{-5pt}
\end{table}

\begin{table}\centering
\small
\begin{tabular}{llccc}
	\toprule
	\textbf{Test set}& \textbf{Model} &  \textbf{B}& \textbf{M} &\textbf{P} \\ 
    \midrule
    \multirow{3}{*}{E2E} & T5-large  &33.23 &35.40 &\textbf{60.18}\\
    & \CC{20} &\CC{20}\textbf{35.51} &\CC{20}\textbf{35.98} &\CC{20}{60.06} \\
    & \CC{20}\multirow{-2}{*}{\modelname} & \CC{20}\posimprov{2.28} &\CC{20}\posimprov{0.58} &\CC{20}\negimprov{-0.12}\\
    
    \midrule
    
    \multirow{3}{*}{DART} & T5-large  &25.94 &33.64 &33.50\\
    & \CC{20} &\CC{20}\textbf{30.42} &\CC{20}\textbf{35.30} &\CC{20}\textbf{36.60} \\
    & \CC{20}\multirow{-2}{*}{\modelname} &\CC{20}\posimprov{4.48}  &\CC{20}\posimprov{1.66} &\CC{20}\posimprov{3.10}\\
	
	\bottomrule
	\end{tabular}
	\vspace{-8pt}
	\caption{Out-of-Domain results. \textbf{B}, \textbf{M} and \textbf{P} represent BLEU, METEOR and PARENT-F1, respectively.}
	\label{table:ood}
	\vspace{-5pt}
	\end{table}

We now assess the model's capability of describing new predicates that are never seen during training. As mentioned in \S\ref{sec:dataset}, WebNLG provides such an official test-unseen set for the evaluation and we construct a similar (but more difficult) test set on DART where all the test predicates are not included in training.
We train the models on WebNLG and DART, and evaluate on the corresponding test-unseen sets, respectively. 
As in \S\ref{sec:exp:few-shot}, we compare \modelname{} with the respective end-to-end T5 models (small, base, large, prefix-tuning).
We also include the previously reported baseline results on the WebNLG test-unseen set, including BestPlan~\citep{moryossef-etal-2019-step}, Pipeline-Trans~\citep{castro-ferreira-etal-2019-neural} and PlanEnc~\citep{zhao2020bridging}.
The experimental results are shown in Table~\ref{table:webnlg_unseen} and Table~\ref{table:dart_unseen}, respectively. 
As can be seen, our method achieves consistent improvements over all the baseline methods, showing the robustness of our method to unseen predicates given the rich commonsense and world knowledge introduced through the pretrained LMs in both stages. 
The superior performance of \modelname{} over the corresponding end-to-end T5 again demonstrates the advantage of our modularization that applies to and improves various pretrained LMs. 
Similar as in the zero-/few-shot experiments, here we observe that on the more difficult DART test-unseen set with more unseen predicates, our method achieves more significant gains than on WebNLG, which further shows the advantage of our method when generalizing to unseen predicates. 

\subsection{Learning with Out-of-Domain Examples} 
At last, we quantitatively measure the generalization ability of our approach across domains.
To simulate the out-of-domain setting, we train our model on the WebNLG dataset and evaluate it on the test sets of DART and E2E, respectively. The DART test set includes the instances from the WebNLG and E2E test sets. We remove those instances to avoid any in-domain test examples (w.r.t the WebNLG training examples) and any overlap with E2E evaluation.
We compare our method with the end-to-end finetuned T5-large model. The experimental results in Table~\ref{table:ood} show that our method outperforms the baseline models on both out-of-domain test sets, echoing the conclusions in previous experiments that our approach with the two-stage design and integration of pretrained LMs has a superior generalization ability to handle data-to-text generation in any-shot scenarios.

\begin{table}[t]\centering\Large
\resizebox{1.05\columnwidth}{!}{
\begin{tabular}{lp{13cm}}
\toprule
     Source & \texttt{<{\color{orange} Zolder, fastest Lap, Liverpool F.C.}> ; <{\color{LimeGreen} Zolder, Date, October 5}>} \\
     Disambig & \texttt{{\color{orange} Liverpool F.C. set the fastest lap in the Zolder.} {\color{LimeGreen} Zolder was on October 5.}}\\
     Fusion & \texttt{{\color{orange} Liverpool F.C. set the fastest lap in the Zolder} {\color{LimeGreen}on October 5.}}\\
     Baseline & \texttt{Zolder's fastest lap is Liverpool F.C. and the date is October 5.} \\
     Human & \texttt{On October 5, 2008, Liverpool F.C. got the fastest lap at a Zolder race.} \\
     \toprule
     Source & \texttt{<{\color{orange} Aleksandra Kovac, associated Band/associated Musical Artist, Bebi Dol}> ; <{\color{LimeGreen} Aleksandra Kovac, associated Band/associated Musical, ArtistK2 Kovac sisters duo}>}\\
     Disambig & \texttt{{\color{orange} Aleksandra Kovac is associated with Bebi Dol.} {\color{LimeGreen} Aleksandra Kovac is associated with K2 Kovac sisters duo.}}\\
     Fusion & \texttt{{\color{orange} Aleksandra Kovac is associated with Bebi Dol} and {\color{LimeGreen} the K2 Kovac sisters duo.}}\\
     Baseline & \texttt{Aleksandra Kovac is an associated band/associated musical artist with Bebi Dol and the K2 Kovac sisters duo.}\\
     Human & \texttt{Aleksandra Kovac is associated with the musical artist Bebi Dol and is part of the band K2 Kovac sisters duo.}\\
\bottomrule
\end{tabular}}
\vspace{-8pt}
\caption{Qualitative examples in the out-of-domain (top) and unseen-predicates (bottom) settings.}
\label{table:case study}
\vspace{-5pt}
\end{table}

\subsection{Case Study}
Table~\ref{table:case study} shows the outputs of our \modelname{} (based on T5-large) after the data disambiguation stage and the sentence fusion stage, on two data in the out-of-domain and unseen-predicates settings, respectively. The  generated words corresponding to different data triples are highlighted in different colors (as in Figure~\ref{fig:model}).
We also provide the results of the T5-large baseline and the human-written references.
As can be seen, \modelname develops a strong generalization ability to out-of-domain data and unseen predicates. In the first example, \modelname successfully disambiguates the triple \texttt{<Zolder, fastest Lap, Liverpool F.C.>} into "\texttt{Liverpool F.C. set the fastest lap in the Zolder}" while the T5 baseline fails to do so and simply generates "\texttt{Zolder's faster lap in Liverpool F.C.}". Also, in the second example, the baseline directly copies "\texttt{associated Band/associated Musical Artist}" in the output while \modelname correctly converts it into "\texttt{is associated with}".

\section{Conclusion}
We have proposed \modelname to deal with the diverse any-shot problems for data-to-text generation. \modelname is composed of two stages, \textit{data disambiguation} that uses prompted GPT-3 to disambiguate input data triples into short sentences, and \textit{sentence fusion} using state-of-the-art pretrained LMs to fuse these sentences into the desired paragraphs. In the process, \modelname integrates rich external implicit knowledge from the large LMs, which ensures strong generalization capability and broad applicability to zero-/few-/full-shot, unseen-predicates, and out-of-domain training scenarios. Extensive experiments show our approach consistently achieves significant improvements over diverse baselines.

\section*{Limitations}
One limitation of our approach is that the data disambiguation stage is done by the GPT-3 model locally, i.e., the GPT-3 model only observes one triple and does not utilize the full-table information. In some difficult cases, the full-table context may be needed for disambiguation. Besides, in this work we directly use the output from GPT-3's as the final disambiguation results, which may be problematic since GPT-3 may not always provide the correct templates, especially when working with highly-specialized domains. In addition, our current approach can only be applied to languages that have access to large LMs. 

\section*{Ethics Statement}
We are aware of the ACL Code of Ethics and the ACM Code of Ethics and Professional Conduct and strictly adhere to the rules throughout the course of this research. 

Our research does not present any new datasets but introduces a new algorithm for data-to-text generation, which generates text descriptions for a given graph or table. The intended usage of the work may potentially provide benefits to people with difficulties in reading graphs or tables, such as people with visual impairment. We do not anticipate direct harm with the intended usage. 

Similar to most generation systems, if harmful input, such as unethical text or input designed for adversarial attacks, exists, our approach is likely to generate unintended output. Therefore, we do not recommend usages of our approach outside controlled research environment before these risks are mitigated. We would also like to point out that a naive deployment of our method may allow malicious exploitation of the backbone Large LMs, thus precautions such as a filtering mechanism need to be implemented.

Our model makes use of the common sense reasoning ability of large LMs, which may reinforce existing social stereotypes, hence care must be taken when applying this approach to materials (e.g. tables and graphs) that are sensitive to populations that already experience marginalization.

Computation-wise, our finetuning procedure takes around 1836 GPU/Hours on NVIDIA GeForce RTX 3090 Ti GPUs. Throughout the study, our prompting module makes about 4600 API calls to Open-AI's GPT-3 API.

\bibliography{anthology,custom}
\bibliographystyle{acl_natbib}

\clearpage


\appendix

\section{GPT-3 Prompt}
\label{sec:prompt}
The prefix in the prompt we use is:
 \\
 \\
\texttt{\footnotesize
Table: Michael | birth Place | USA\\
Text: Michael was born in the USA.\\
 \\
Table: First Clearing | location | On NYS 52 1 Mi. Youngsville\\
Text: First Clearing is located at On NYS 52 1 Mi. Youngsville.\\
 \\
Table: Abilene Regional Airport | city Served | Abilene Texas\\
Text: Abilene Regional Airport serves Abilene Texas.\\
 \\
Table: Alfred Moore Scales | active Years Start Date | 1875-03-04\\
Text: Alfred Moore Scales started to be active on 1875-03-04.}
\\

\section{Experimental Details}
We use a batch size of 5 and a beam search size of 5 for zero-shot and few-shot settings. For other settings, we do model selection based on the performance on the validation set, with a batch size chosen from \{2, 4, 8\} and \{1, 3, 5\}, respectively. We use sacreBLEU~\citep{post-2018-call} for model selection. The URL for the metrics and corpus we use are shown in Table~\ref{table:metric} and Table~\ref{table:dataset}, respectively.

\begin{table}[h]\centering
\begin{tabular}{lp{5cm}}
\toprule
    Metric & URL \\
    \midrule
     BLEU & \url{https://github.com/moses-smt/mosesdecoder/blob/master/scripts/generic/multi-bleu.perl} \\
     \midrule
     METEOR & \url{https://www.cs.cmu.edu/~alavie/METEOR/index.html} \\
     \midrule
     PARENT & \url{https://github.com/KaijuML/parent} \\
     \midrule
     BERTScore & \url{https://github.com/Tiiiger/bert_score} \\
     \midrule
     BLEURT & \url{https://github.com/google-research/bleurt} \\
     \midrule
     SacreBLEU & \url{https://github.com/mjpost/sacrebleu}\\
\bottomrule
\end{tabular}
\caption{The URLs for the metrics we use in the experiments.}
\label{table:metric}
\end{table}

\begin{table}[h]\centering
\begin{tabular}{lp{5cm}}
\toprule
    Dataset & URL \\
    \midrule
     WebNLG & \url{https://gitlab.com/shimorina/webnlg-dataset/-/tree/master/webnlg_challenge_2017} \\
     \midrule
     DART & \url{https://github.com/Yale-LILY/dart} \\
     \midrule
     E2E & \url{https://github.com/tuetschek/e2e-dataset} \\
     \midrule 
     WikiFluent & \url{https://github.com/kasnerz/zeroshot-d2t-pipeline}\\
\bottomrule
\end{tabular}
\caption{The URLs for the corpus we use in the experiments.}
\label{table:dataset}
\end{table}

\section{Zero-/Few-shot Experimental Results}\label{sec:appendix any shot}

\begin{table*}\centering
\small
\begin{tabular}[t]{lccccc}
\toprule
\#Instance  & 0 & 10                         & 20                         & 50                         & 100                        \\ \midrule
KGPT     &   14.19/20.78/20.67  & 17.50/23.13/ 25.77          & 18.40/23.44/26.49          & 21.68/25.30/29.22          & 24.72/26.71/46.50          \\
T5-large   &  10.46/25.63/23.67 & 24.74/32.28/42.48          & 41.38/36.12/52.77          & 45.32/39.49/59.39          & 48.68/39.24/60.66          \\      
\midrule
\modelname{} & \textbf{43.99/39.32}/58.23 & \textbf{45.16/38.95}/58.24 & \textbf{47.46/39.35}/59.85 & \textbf{49.36/40.08}/61.25 & \textbf{49.39/40.09}/61.08 \\ 
\ \emph{- w/o weak-sup} & 34.47/30.06/51.51 & 39.38/33.93/56.44 & 43.67/35.81/57.99 & 47.56/38.61/60.04 & 48.60/39.68/60.56\\
\ \emph{- w/ manual templ.} & 42.02/38.85/\textbf{58.26} & 43.37/38.69/\textbf{58.80} & 46.12/38.88/\textbf{60.94} & 48.28/39.64/\textbf{62.02} & 48.32/39.32/\textbf{61.92} \\
\bottomrule
\end{tabular}\caption{WebNLG few-shot results. $x$ / $y$ / $z$ denotes the model performance on BLEU / METEOR / PARENT-F1.}
\label{table:webnlg any shot}
\end{table*}

\begin{table*}\centering
\small
\begin{tabular}[t]{lccccc}
\toprule
\#Instance  & 0 & 10                         & 20                         & 50                         & 100                        \\ \midrule
KGPT     &   85.35/43.78  & 88.62/49.67         & 88.92/49.41          & 89.66/52.72          & 90.30/55.15          \\
T5-large   & 84.17/40.19  &  93.00/67.49 & 92.87/65.93    & 93.06/66.48          & 93.24/67.05          \\      
\midrule

\modelname{} & \textbf{92.43/71.93} & \textbf{94.39/72.45} & \textbf{ 94.69/73.48 } & \textbf{ 95.03/74.62 } & \textbf{ 94.99/74.66 } \\ 
\ \emph{- w/o weak-sup} & 92.01/66.43 & 93.05/67.84  & 93.10/67.32 & 93.54/68.10 & 93.93/68.05 \\
\ \emph{- w/ manual templ.} & 92.36/71.01 & 94.17/72.08 & 94.27/72.91 & 94.58/74.11 & 94.61/74.33 \\
\bottomrule
\end{tabular}\caption{WebNLG few-shot results. $x$ / $y$ denotes the model performance on BERTScore / BLEURT.}
\label{table:webnlg any shot 2}
\end{table*}

\begin{table*}[t]\centering
\small
\begin{tabular}{lccccc}
\toprule
\#Instance  & 0  & 10                         & 20                         & 50                         & 100                        \\ \midrule
KGPT      &  11.15/19.30/18.92  & 14.91/19.74/23.76           & 16.83/21.30/26.67          & 20.16/23.14/31.13         & 20.31/23.82/31.35          \\
T5-large  &  8.43/22.67/23.81  & 29.97/31.44/46.82          & 32.96/31.76/47.36          & 37.08/34.43/54.10          & 39.92/34.90/55.05         
     \\
\midrule 
\modelname  & \textbf{38.81/36.91/54.10} & \textbf{40.50/36.65/56.00} & \textbf{41.45/36.45/57.34} & \textbf{42.33/36.99/57.63} & \textbf{42.87/36.77/58.37} \\ 
\ \emph{- w/o weak-sup} & 31.92/26.15/43.99 & 38.15/32.11/54.97          & 37.12/32.80/54.12          & 40.79/35.70/56.40          & 41.22/35.15/57.79 \\
\bottomrule
\end{tabular}\caption{DART few-shot results. $x$ / $y$ / $z$ denotes the model performance on BLEU / METEOR / PARENT-F1.}
\label{table:dart any shot}
\end{table*}

\begin{table*}\centering
\small
\begin{tabular}[t]{lccccc}
\toprule
\#Instance  & 0 & 10                         & 20                         & 50                         & 100                        \\ \midrule
KGPT     &   84.32/43.21  & 87.13/48.94         & 88.54/49.22          & 89.43/52.13          & 89.96/53.99          \\
T5-large   & 83.53/40.01  &  88.73/66.37 & 89.43/66.65    & 90.39/66.79         & 90.51/66.96          \\      
\midrule

\modelname{} & \textbf{90.13/69.87} & \textbf{91.52/69.88} & \textbf{ 91.67/70.10 } & \textbf{ 91.90/70.46 } & \textbf{ 92.01/70.61 } \\ 
\ \emph{- w/o weak-sup} & 88.94/67.96 & 90.21/68.13  & 90.44/68.37 & 90.56/68.46 & 90.84/68.66 \\
\bottomrule
\end{tabular}\caption{DART few-shot results. $x$ / $y$ denotes the model performance on BERTScore / BLEURT.}
\label{table:dart any shot 2}
\end{table*}

We show the BLEU/METEOR/PARENT-F1 scores for zero-/few-shot experiments on WebNLG and DART in Table~\ref{table:webnlg any shot} and Table~\ref{table:dart any shot}, and 
{BERTScore/BLEURT in Table~\ref{table:webnlg any shot 2} and Table~\ref{table:dart any shot 2}}.

\end{document}